\RequirePackage{amsmath}
\documentclass[runningheads]{llncs}
\usepackage{blindtext}
\usepackage{cite}
\usepackage{amsmath,amssymb,amsfonts}
\usepackage{enumerate}
\usepackage{footnote}
\usepackage{url}
\usepackage{algorithmic}
\usepackage{graphicx}
\usepackage{float}
\usepackage{textcomp}
\usepackage{comment}
\usepackage{color,soul}
\usepackage{amsfonts}
\usepackage{xcolor}
\usepackage{hyperref}

\usepackage[font=small,skip=0pt]{caption}
\usepackage{tablefootnote}

\def\BibTeX{{\rm B\kern-.05em{\sc i\kern-.025em b}\kern-.08em
    T\kern-.1667em\lower.7ex\hbox{E}\kern-.125emX}}
    
\begin{document}

\title{NatCSNN: A Convolutional Spiking Neural Network for recognition of objects extracted from natural images}

\author{Pedro Machado\inst{1}\orcidID{0000-0003-1760-3871} \and
Georgina Cosma\inst{1}\orcidID{0000-0002-4663-6907} \and
T.M. McGinnity\inst{1}\orcidID{0000-0002-9897-4748}}
\authorrunning{Machado \textit{et al.}}
\titlerunning{NatCSNN for recognition of objects...}
%
\institute{\textit{Computational Neurosciences and Cognitive Robotics Group} \\
\textit{School of Science and Technology}\\
\textit{Nottingham Trent University} \\
Nottingham, United Kingdom\\
\email{\{pedro.baptistamachado,georgina.cosma,martin.mcginnity\}@ntu.ac.uk}}

\maketitle

\begin{abstract}
Biological image processing is performed by complex neural networks composed of thousands of neurons interconnected via thousands of synapses, some of which are excitatory and others inhibitory. Spiking neural models are distinguished from classical neurons by being biological plausible and exhibiting the same dynamics as those observed in biological neurons. This paper proposes a Natural Convolutional Neural Network (NatCSNN) which is a 3-layer bio-inspired Convolutional Spiking Neural Network (CSNN), for classifying objects extracted from natural images. A two-stage training algorithm is proposed using unsupervised Spike Timing Dependent Plasticity (STDP) learning (phase 1) and ReSuMe supervised learning (phase 2). The NatCSNN was trained and tested on the CIFAR-10 dataset and achieved an average testing accuracy of 84.7\%  which is an improvement over the 2-layer neural networks previously applied to this dataset.

\end{abstract}

\section{Introduction} \label{introduction}
The mammalian visual cortex is responsible for performing advanced, complex and low-power (about 20 watts \cite{Ling2001}) image processing. Neuromorhpic architectures (neuro-biological architectures that can run bio-inspired models of neural systems) have evolved as a consequence of the rapid miniaturisation of electronic components, lithography manufacturing process and the developments of cognitive applications \cite{Chen2018}. In particular, Spiking Neural Networks (SNN), which are characterised by displaying similar spike-timing encoding and plasticity as real neurons \cite{Izhikevich2004}, offer a sophisticated low-power and high performance computational processing paradigm. However, the use of SNN for classification of natural images with a high accuracy remains a complex task on typical machine/deep learning benchmarks such as CIFAR-10 \cite{Krizhevsky2009}.

This paper proposes NatCSNN, a bio-inspired Convolutional Spiking Neural Network for natural image object classification. The proposed architecture, includes a two phase training approach where: in phase 1, unsupervised Spike Timing Dependent Plasticity (STDP) learning is used for training the middle layers; and in phase 2, the ReSuMe supervised learning algorithm is used to train the Layer 3 neurons. A systematic method for searching for the initial synaptic weights is also proposed. 

The paper is structured as follows: the literature review is discussed in section~\ref{lit_review}, the NatCSNN architecture is discussed in section~\ref{architecture}, the experimental training of the NatCSNN is discussed in section~\ref{methodology}, the results are shown in section~\ref{results} and a discussion and future work are provided in section~\ref{discussion}.

\section{Literature review} \label{lit_review}
The majority of previous works have focused on conventional CSNN and the MNIST dataset. Wang \textit{et al.} \cite{Wang2018}, proposed a similarity search method using a forward SNN with successively connected encoding. The authors claim an accuracy of 100$\%$, 100$\%$ and 92$\%$ for noise levels of 5$\%$, 20$\%$ and 40$\%$ respectively when tested on the MNIST dataset \cite{lecun-mnisthandwrittendigit-2010}. The authors proposed a new method for training multi-layer spiking convolution neural networks (CSNN) incorporating supervised and unsupervised learning. The training process includes two components for unsupervised feature extraction and supervised classification using adapted versions of the Spike-Timing Dependent Plasticity (STDP). Tavanaei \textit{et al.} \cite{Tavanaei2018} claim that their proposed CSNN achieved an accuracy of 98.60$\%$ on the MNIST dataset \cite{lecun-mnisthandwrittendigit-2010}. Kheradpisheh \textit{et al.} \cite{Kheradpisheh2018}, designed an STDP-based spiking deep CSNN composed of one Difference of Gaussians (DoG) layer (temporal encoding) three convolutional layers and three pooling layers, using unsupervised STDP learning. The paper proposes an eight layer architecture SNN and claim an accuracy of 98.4$\%$ when tested in the MNIST dataset. Kulkarni \textit{et al.} \cite{Kulkarni2018}, proposed a three layer architecture trained with a supervised learning approach using the spike triggered Normalised Approximate Descendent algorithm with an accuracy of 98.17$\%$ on the MNIST dataset. In \cite{Lee2018}, a deep spiking CSNN (SpiCNN) composed of a hierarchy of stacked convolution layers, a spatial-pooling layer and a fully-connected layer is proposed. The SpiCNN was trained using unsupervised STDP with an accuracy of 91.1$\%$ on the MNIST dataset. 

Relevant literature on CSNNs \cite{Wang2018, Tavanaei2018, Kheradpisheh2018, Kulkarni2018, Lee2018} for image processing have in common: (i) multiple-layer CSNNs, (ii) use of the Leaky-integrate-and-fire (LIF) neuron model, (iii) use of unsupervised/supervised STDP (or an STDP adaptation) for training and (iv) all of the relevant work was tested on the MNIST hand-written black-and-white dataset.

The MNIST dataset is one of the commonly used benchmark datasets, composed only of simple black-and-white images (containing hand-written numbers from 0 to 9). In contrast, the CIFAR-10 \cite{Krizhevsky2009} dataset is composed of 50,000 coloured natural images with natural backgrounds and thus represents a sigificantly more challenging test for image processing algorithms. 

More recent works, Sengupta \textit{et al.} \cite{Sengupta2018} and Hu \textit{et al.} \cite{Hu2018} have proposed hybrid architectures by combining classical deep learning architectures with spiking neural networks. Sengupta \textit{et al.} \cite{Sengupta2018} proposed an ANN to SNN conversion technique that is claimed to outperform state-of-the-art techniques, reporting an $12.54\%$ error on the CIFAR-10 dataset when combined with Residual network architectures. Hu \textit{et al.} \cite{Hu2018}, proposed a shortcut normalisation mechanism to convert continued-valued activation's to match firing rates in SNN; Their proposed architecture receives the continued-valued activation's from a Residual Network and converts them into spiking rates which are fed into a spiking residual network architecture. Their experiemnts achieved an accuracy of $92.85\%$ when tested on the CIFAR-10 dataset. Despite the accuracy obtained on the CIFAR-10 using these hybrid architectures, the proposed architectures are all composed of several layers and none of the architectures uses biological spiking neuron parameters, reducing therefore the bio-plausibility of the proposed architectures.

Other work published on the CIFAR-10 has utilised classical deep neural networks. Krizhevsky \textit{et. al.} \cite{Krizhevsky2010} describe how to train a two-layer convolutional Deep Belief Network(DBN) on the CIFAR-10 and obtained 78.90$\%$ accuracy using a 2 layer architecture similar to the NatCSNN proposed in this paper. In \cite{Srivastava2015}, a 32-layer network, designated as highway networks inspired in Long Short-Term Memory (LSTM), is proposed and the authors reported an accuracy of 92.40$\%$ on the CIFAR-10 dataset. Springenberg \textit{et al.} \cite{Springenberg2014}, proposed a 10-layer Network, designated as All-CNN, with an accuracy of 92.40$\%$ without data augmentation on the CIFAR-10 dataset. 

The NatCSNN proposed in this paper is a compact, low-layer count (3) bio-inspired architecture, target at processing natural images where the network must perform the task of extracting features from CIFAR-10. It is implemented as a multi-hierarchical SNN composed of three SNN layers connected via excitatory and inhibitory synapses and trained using unsupervised STDP learning (layers 1 and 2) and supervised learning using ReSuMe (details are presented in section \ref{architecture}). This paper also proposes a systematic method to search for the initial synaptic weights and a 2-phase training approach, using a mixture of unsupervised and supervised learning. NatCSNN was compared to two adaptations of the architecture proposed in \cite{Krizhevsky2009}. 

\section{NatCSNN architecture} \label{architecture}
\subsection{Spiking Neuron Model} \label{neurons}
\begin{figure} [h] \small
	\begin{center}
	\includegraphics[width=0.3\textwidth]{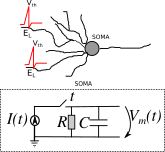}
	\end{center}
	\caption{Equivalent electronic circuit of the leaky-integrate-and-fire neuron model. Let a current $I(t)$ charge the $RC$ circuit. Spikes are generated when the voltage $V_m(t)$ at the terminals of the capacitance is greater or equal to the threshold $V_{th}$ making $V_m(t)$ dropping to $E_L$ (reset voltage) during a $t_{ref}$ (refractory period). \label{fig:LIF}} 
\vspace{-4mm}
\end{figure}

The neuron model used in this work is based on the Leaky-Integrate-and-Fire (LIF) neuron model was selected because it is considered to be one of the simplest spiking neuron models describing the biological neuronal cells dynamics \cite{Gerstner2014}. More complex models with higher computational requirements are available (\textit{e.g.} Izhikevich \cite{Izhikevich2004}) but demand higher computational requirements. The LIF dynamics can be represented as a RC (Resistance, Capacitance) electronic circuit as depicted in picture Figure~\ref{fig:LIF}.

The LIF neuron model is governed by the equation \ref{eq:1}.
\begin{eqnarray}
\label{eq:1}
\tau_m\frac{\delta V_m}{\delta t} =- V_m + RI \left( t \right)\\
\nonumber
\end{eqnarray}
\vspace{-4mm}

where $\tau_m = RC$ is the time constant, $R$ the membrane resistance, $C$ the membrane capacitance, $V_m(t)$ the membrane voltage and $I(t)$ is the current at time $t$.
When $V_m(t)$ reaches the $V_{th}$ (threshold voltage), the membrane voltage is set to the reset membrane potential ($V_m(t)=E_L$). 

Equation~\ref{eq:1} can be improved using the multi-timescale adaptive threshold predictor and non-resetting leaky integrator (MAT) as described in \cite{Kobayashi2009}. MAT provides an adaptive threshold and prevents the neuron from over-spiking when exposed to high continuous currents. The adaptive threshold is described by equations~\ref{eq:2} and \ref{eq:3} and taken from the implemented version of the MAT in the NEST framework\cite{NEST2018}.
\begin{align}
\label{eq:2}
  V_{th}(t)=\sum_k H(t-t_k) + E_L
\end{align}
where 
\begin{align} 
\label{eq:3}
H(t)=\sum_{j=1}^{L}w_j^{(-t \tau_{mj})}
\end{align}

where $V_{th}$(t) is the threshold voltage in time $t$, $t_k$ is the $k^{th}$ spike time, $L$ is the number of threshold time constants, $\tau_{mj}$ $(j=1,...,L)$ are the $j^{th}$ time constants, $w_j$ $(j=1,...,L)$ are the weights of the $j^{th}$ time constants, and $E_L$ is the reset membrane potential value \cite{Kobayashi2009}.

The MAT model, implemented in NEST\footnote{retrieved from \protect\url{https://nest-simulator.readthedocs.io/en/latest/models/neurons/integrate_and_fire/iaf_psc_alpha.html?highlight=iaf_psc_alpha}, last accessed on the 26/03/2019}, was used in our work as the neuron model for providing the desired adaptive threshold dynamics.

\subsection{Layers and synaptic connectivity}
\begin{figure}[h] \small
	\begin{center}
	\includegraphics[width=.45\textwidth]{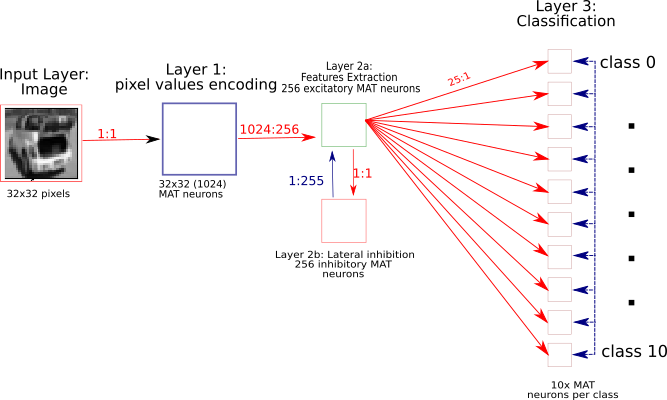}
	\end{center}
	\caption{NatCSNN with (i) n$\times$m (n and m $\in \mathbb{Z}$) (image input followed by the three processing layers. Layer 1: Encoding of pixel intensity values to spike events, Layer 2a: features extraction, Layer 2b: lateral inhibition and Layer 3: classification layer. \label{fig:proposed_architecture}} 
	\vspace{-4mm}
\end{figure}

The input layer receives an image of n rows $\times$ m columns (n and m $\in \mathbb{Z}$); Layer 1 performs the encoding of the pixel intensity values to spike events, Layer 2a extracts features, Layer 2b provides lateral inhibition, Layer 3 classifies the object type. The proposed architecture is shown in Figure~\ref{fig:proposed_architecture}. 

\textbf{Input images to the NatCSNN} 

Each image is converted into grayscale, normalised (pixel intensity values in the range of 0.0 up to 1.0) and its pixel intensity value converted into spike train events. The image pixel intensity values are feed to the Layer 1 neurons using a one-to-one (1:1) connectivity. Each pixel intensity value is converted into a current given by equation~\ref{eq:4} 
\vspace{-0mm}
\begin{align}
\label{eq:4}
I(i,j) = p(i,j).I_K 
\vspace{-6mm}
\end{align}

where I(i,j) is the current for a the neuron in row i and column j, p(i,j) is the pixel intensity value in row i and column j and $I_K$ is a current constant for producing the desirable spike rate. 

\subsubsection{Layer 1: Pixel intensity values encoding}

Each neuron, in Layer 1, receives a current proportional to a pixel intensity value given by equation~\ref{eq:4}. The frequency of spikes is linearly proportional to the pixel intensity value and therefore a neuron will spike up to 10 times during a simulation time step if the pixel intensity value is near the maximum (1.0) or none if near the minimum value (0.0). The neurons in Layer 1 connect to the neurons in Layer 2a (features extraction group) via all-to-all connections.

STDP synapses are used during phase 1 training to adjust the weights of the connections between the pre-neurons (Layer 1) to the post-neurons (Layer 2a). The weights change based on the STDP parameters (detailed in section~\ref{methodology}) and trained on the training dataset of 50,000 images during 5 epochs (runs).

\textbf{Layer 2: Feature extraction} 

Layer 2 is composed of 2 groups of neurons, one called the features extraction group and the other the lateral inhibition group. Layer 2a and Layer 2b have the same number of neurons, which is 25$\%$ of the Layer 1 neurons. Each neuron in the feature extraction group receives input from all Layer 1 neurons. Each neuron in the feature extraction group presents its output to a single neuron of the lateral inhibition group via one excitatory synapse and receives incoming spike events from all the other neurons in the lateral inhibition group. 

STDP synapses are used during phase 1 of training to connect the Layer 2a neurons (features extraction group) to the Layer 2b neurons (lateral inhibition group) and vice-versa. The number of neurons in both Layer 2a and 2b is the same. The weights of Layer 2a and 2b neuron's synapses change accordingly to the STDP parameters (detailed in section~\ref{methodology}) when trained on the training dataset of 50,000 images during 5 epochs. The number of epochs was experimentally obtained and 5 epochs was the value that produced better results. Values above 5 epochs made the synaptic weights to saturate in the maximum weights allowed for each synapse and values below 5 epochs produced lower accuracy.

\textbf{Layer 3: Classification}
Layer 3 is composed of 10 groups (one group per class) of 10 LIF neurons. Each neuron of the Layer 2a feature extraction group is connected to all the Layer 3 neurons. The 10 LIF neurons per class are used to improve the classification accuracy by increasing the resolution. 

Layer 3 neurons are only trained in phase 2 training after training the Layer 2 neurons (phase 1 training) during the initial 5 epochs. During phase 2 training, the synapses of the Layer 2 neurons are converted into static synapses with fixed weights where the final Phase 1 trained weights are used. During phase 2, the neurons of the feature extraction group are connected using STDP synapses and their weights are trained on the training dataset during 5 further epochs. As before, the 5 epochs were selected experimentally.

\begin{figure} [h]
	\begin{center}
	\includegraphics[width=0.5\textwidth]{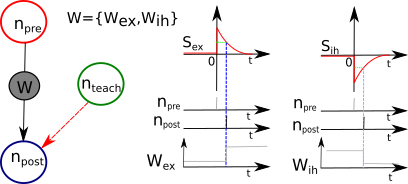}	
	\end{center}
	\caption{ReSuMe learning: (A) Remote supervision. (B) Learning windows \protect\cite{Ponulak2010}. \label{fig:resume}}
	\vspace{-4mm}
\end{figure}

ReSuMe \cite{Ponulak2010}, a supervised learning algorithm, was used for training the response of the Layer 3 neurons. In ReSuMe, teacher signals are used for producing the desired spike pattern in response to a stimulus \cite{Ponulak2010}. 
Figure~\ref{fig:resume} shows the teacher signal (desired spike pattern) $n_{teach}$ being presented to a neuron $n_{post}$ for delivering a spike pattern by adjusting the synaptic weight w between the pre-neuron $n_{pre}$ and the post-neuron $n_{post}$. The learning occurs with modification of the weights.
 
\vspace{-10mm}
\newpage
The ReSuMe \cite{Ponulak2010} equations are as follows:
\begin{align} [h]
\label{eq:5}
W_{ex}(s_{ex})=
  \begin{cases}
  A_{ex} e^{\left( \frac{-s_{ex}}{\tau_{ex}}\right)}, & \text{if }s_{ex}>0,\\
  0, & \text{if } s_{ex} \leq 0,
  \end{cases}\\
W_{ih}(s_{ih})=
  \begin{cases}
  A_{ih} e^{\left( \frac{-s_{ih}}{\tau_{ih}}\right)}, & \text{if }s_{ih}>0,\\
  0, & \text{if } s_{ih} \leq 0,
  \end{cases}
  \vspace{-4mm}
\end{align}

where $A_{ex}$, $A_{ih}$, $\tau_{ex}$ and $\tau_{ih}$ are constants. $A_{ex}$ and $A_{ih}$ are positive in excitatory synapses and negative in inhibitory synapses. In both cases $\tau_{ex}$ and $\tau_{ih}$ are positive time constants \cite{Ponulak2010}.

\section{Training the NatCSNN} \label{methodology}
This section specifies the methodology that was followed for training the NatCSNN network and evaluating its performance on the CIFAR-10 dataset \cite{Krizhevsky2009}. 
The simulation was performed using the NEST (NEural Simulator Tool) version 2.16.0 \cite{NEST2018}.

\subsection{Building the NatCSNN}
The simulation setup included the following steps:

\textbf{Step 1 - Load dataset to memory:} The dataset is loaded to memory and the images converted to grey-scale. The conversion of the images into grey-scale reduces substantially (a third) the required number of neurons and synapses. Grey-scale images were used for reducing the number of spiking neurons and its synapses.

\textbf{Step 2 - Convert pixel intensity values to currents:} Pixel intensity values have to be multiplied by a current constant $I_k$ to get a spike pattern proportional to the pixel intensity value and the spikes are regularly spaced during the period of 100ms (simulation time-step). $I_k$ was modelled so that a given neuron spikes up to 10 times over a period of 100ms.

\textbf{Step 3 - Create the network:} The NatCSNN architecture is created as follows: Layer~1: 1024 neurons (32 rows $\times$ 32 columns), Layer~2: 512 neurons (1024 $\div$ 4 $\times$ 2 groups), Layer~3:  100 neurons (10 classes $\times$ 10 neurons).

\textbf{Step 4 - Create synaptic connectivity using:} L1 to L2 are connected via all-to-all connectivity, L2a (features extraction group) to L2b (lateral inhibition group) via one-to-one connectivity, L2b to L2a via one-to-(n-1) (all the neurons with exception of the neuron that is connected to the neuron in L2a) and L2a to L3 via all-to-all connectivity.

\textbf{Step 5 - Connect the L2a to L3 neurons:} The L2 neurons are connected to the L3 neurons via all-to-all connections. Each L3 neuron of each class receives connection of one 10$\%$ of the L2a neurons). 
Overall, the 10 neurons per classifier receives outputs of of the L2b neurons. Each L3 neurons of a given class connect to all the other L3 neurons classes via inhibitory synapses.

\textbf{Step 6 - Set the simulation parameters:} The simulation is configured with a time step of $t=100 ms$ and the neurons with the parameters are as follows: initial $V_m$=-70.0 mV, $E_L$ = -70.0 mV, $C_m$ = 100.0 pF, $\tau_m$ = 5.0 ms, $\tau_{syn_{ex}}$ = 1.0 mV, $\tau_{syn_{in}}$ = 3.0 ms, $t_{ref}$ = 2.0 ms, $t_{spike}$ = -1.0 ms, $\tau_1$ = 10.0 ms, $\tau_2$=20.0 ms, $\alpha_1$ = 37.0 mV, $\alpha_2$=2.0 mV, $\omega$ = -51.0 mV, $V_{th}$ = -51.0 mV and $V_{reset}$ = -70.0 mV. 

\textbf{\textit{Training phase 1:}} During phase 1 training the synapses of L1 to L2a, L2a to L2b and L2b to L2a are trained using unsupervised STDP with the parameters listed in table~\ref{tab:STDP_ReSuMe}. The STDP parameterisation was selected from the parameters suggested by Gerstner \textit{et al.}  \cite{Gerstner2014} that have been observed in the Visual Cortex and Hippocampal.

The 50,000 training images of the training batch were presented, one-by-one, for a period of 100ms (one simulation timestep) to the network in 5 epochs and during the phase 1 of training the weights were adjusted accordingly to the STDP rules. The weights were stored, into files, every 500 simulation time steps (or 50000 ms of simulation).

\textbf{\textit{Training phase 2:}}
The synapses of L1 to L2a, L2a to L2b and L2b to L2a are converted to static synapses using the weights trained in phase 1. The excitatory and inhibitory synapses are of STDP type. During Phase 2 training, the weights of the neurons in Layer 3 STDP synapses are trained using the ReSuMe algorithm and parameters were set as listed in table~\ref{tab:STDP_ReSuMe}:
\begin{table} [h] 
 \caption{Unsupervised STDP and ReSuMe parameters}
 \label{tab:STDP_ReSuMe}
 \begin{center}
  \begin{tabular}{| c | l | c | c |} 
 \hline
 Parameter & Description & unsupervised STDP & ReSuMe \\ 
 \hline\hline
 $W_{ex}$ &  initial excitatory weight & random($600.0\pm 10\%$)\tablefootnote{L1 to L2a neurons} & 241.\\
  & excitatory synapse &  random$(490.84\pm 10\%$)\tablefootnote{from L2a to L2b neurons} &\\
 \hline
 $W_{ih}$ & initial inhibitory weight & random$(-100.0\pm 10\%$) & -120. \\
 & inhibitory synapse & &\\
 \hline
 $\tau_{ex}$ & time constant of short pre-synaptic trace & 10. ms  & 10.0 ms\\
 \hline
 $A_{ex}^+$ & weight of pair potentiation rule of the & 0.001 & 0.001\\
 & excitatory synapse & &\\
 \hline
 $A_{ex}^-$ & weight of pair depression rule of the & 0.0005 &  0.0 \\
  & excitatory synapse & &\\
 \hline
 $A_{ih}^+$ & weight of pair potentiation rule of the & 0.001 & 0.001\\
  & inhibitory synapse & &\\
 \hline
 $A_{ih}^-$ & weight of pair depression rule of the & 0.0005 & 0.0 \\
  & inhibitory synapse & &\\
 \hline
 $Wmax_{ex}$ & maximum allowed weight of the & 1200. & 1200.\\
 & excitatory synapse & &\\
 \hline
 $Wmax_{ih}$ & maximum allowed weight of the & -1200. & 1200.\\
 & inhibitory synapse & &\\
 \hline
 \hline
\end{tabular}
\end{center}
\vspace{-8mm}
\end{table}

Ten extra neurons (one per class) are used to provide the teaching signals to the classifier neurons as specified using the ReSuMe algorithm. Each teaching neuron will only spike when a picture being exposed to the neurons in Layer 1 belongs to that class. The 50,000 training images of the training batch were presented to the network during 5 epochs and during that period the weights were adjusted accordingly to the STDP rules and to the teaching signals applied by the teaching neurons to the Layer 3 classifier neurons.

\textbf{\textit{Testing mode:}}
All the STDP synapses were replaced by static synapses using the weights trained in phases 1 and 2.

\section{Results} \label{results}
The NatCSNN was trained using the 50,000 testing images in two phases, phase 1, training of the Layer 2 neurons synapses, using unsupervised STDP learning for 5 epochs and phase 2, training of the excitatory and inhibitory synapses of the layer 3 neurons, using supervised learning using ReSuMe for 5 epochs. 

The remaining 10,000 CIFAR-10 images were reserved for testing. One of the most challenging tasks in spiking neural network simulations is the definition of the starting weights for the network.  In this case we have used the Monte Carlo algorithm to select the initial weight values for the STDP synapses between the L2a features extraction group and the L3 neurons. The initial values of the L2a to L3 synapses is crucial because the selection of a low or high value will impact on the overall accuracy.
The initial value was selected using the Monte Carlo algorithm. The accuracy of the NatCSNN was improved by using 100 neurons (10 per classifier) and using the connectivity shown in Table~\ref{tab:accuracy}. The average value of the obtained accuracy using the 10 neurons per classifier (100 neurons) was 84.70$\%$ with a standard deviation of 1.579$\%$.

All neurons' action potentials were reset before exposing the next image to the NatCSNN. Forcing the neurons to start with the reset voltage is necessary to prevent receipt of an inhibitory stimulus (\textit{i.e.} neurons from Layer 2a and Layer 3), which would cause a drop in membrane action potential to a very low value preventing neurons from spiking when exposed to the excitatory stimulus. The neuron parameters are the same for all the neurons. The initial minimum weight of the STDP excitatory synapses was selected experimentally. Neurons with all-to-all connectivity receive spike contributions from many neurons and therefore the initial weight must below the current required to make the neuron spike (see table~\ref{tab:STDP_ReSuMe}) and below the maximum value (1200.0 $\mu A$, see Table~\ref{tab:STDP_ReSuMe}) that makes neurons spike constantly.

Table~\ref{tab:accuracy} compares the performance of NatCSNN with other relevant approaches. The works in \cite{Springenberg2014}, \cite{Coates2011} and \cite{Krizhevsky2010} have tested their methods on the colour images while the NatCSNN was tested on grey-scale images. To enable a grey-scale comparison, we have re-implemented CDBN \cite{Krizhevsky2010} that utilise grey-scale images and applied it to the CIFAR-10 dataset, because it has the same number of layers as NatCSNN, resulting in two variants (a) CDBN-ANN 1 with two convolutional layers followed by a dense layer with 10 neurons and CDBN-ANN 2 with three convolutional layers followed by a dense layer of 10 neurons. Both CDBN-ANN 1 and CDBN-ANN 2 receive grey images. For a fair comparison the CDBN was tuned to achieve its highest accuracy. Both the CDBN-ANN 1 and CDBN-ANN2 implementations were trained with 100 epochs.
\vspace{-10mm}
\begin{table} [h] \small
 \caption{Classification accuracy of the NatCSNN compared with other classical CNNs tested on the CIFAR-10 dataset}
 \label{tab:accuracy}
 \begin{center}
 \begin{tabular}{| c | c | c | c |} 
 \hline
 Architecture & CIFAR-10 accuracy [$\%$] & Number of layers & images type \\
 \hline\hline
 All-CNN \cite{Springenberg2014} &  92.75 & 10 & Colour \\
 \hline
 Highway Network \cite{Coates2011} & 92.40 & 32 & Colour \\
  \hline
 CDBN \cite{Krizhevsky2010} & 78.90 & 3 & Colour \\
 \hline
 \textbf{NatCSNN (grey)} & \textbf{84.70} & \textbf{3} & \textbf{Grey}\\
  \hline
 CDBN-ANN 1 (grey) & 80.54 & 3 & Grey\\
  \hline
 CDBN-ANN 2 (grey) & 82.4 & 4 & Grey \\
  \hline
 \hline
\end{tabular}
\end{center}
\vspace{-6mm}
\end{table}
\vspace{-10mm}
\begin{table} [h] 
 \caption{Classification accuracy of the NatCSNN using 10 neurons per classifier }
 \label{tab:accuracy_per_class}
 \begin{center}
 \begin{tabular}{| c | c | c | c | } 
 \hline
 Class & NatCSNN [$\%$] & CDBN-ANN 1 [$\%$] & CDBN-ANN 2 [$\%$] \\ 
 \hline\hline
 airplane &  84.005 & 83  & 84 \\
 \hline
 automobile & 87.021 & 92 & 91 \\
  \hline
 bird & 86.119 & 75 & 74 \\
 \hline
 cat & 85.3 & 67 & 66 \\
  \hline
 deer & 83.436 & 78 & 81 \\
  \hline
 dog & 83.421 & 74 & 78 \\
  \hline
 frog & 86.732 & 73 & 81 \\
  \hline
 horse & 82.502 & 88 & 88 \\
 \hline
 ship & 83.35 & 88 & 88 \\
 \hline
 truck & 85.115 & 87 & 89 \\
  \hline
 \hline
\end{tabular}
\end{center}
\vspace{-10mm}
\end{table}
Table~\ref{tab:accuracy} shows that the networks with better accuracy are the ones with more layers and that the NatCSNN has a better accuracy when compared with the colour CDBN \cite{Krizhevsky2010}, the CDBN-ANN 1 (re-implemented by the authors) and the CDBN-ANN 2 (re-implemented by the authors).
It can be seen that the number of layers and the conversion of colour to grey-scale images negatively affects the classification accuracy. Therefore, it will be possible to improve the accuracy of the NatCSNN by adding more layers, although this would increase the complexity of training with the increase of neuron and its synaptic connectivity. We also note that Table~\ref{tab:accuracy_per_class} illustrates a more uniform level of accuracy of NatCSNN across classes, as compared to CDBN.

\section{Discussion and Future Work} \label{discussion}
This paper proposes, NatCSNN, a bio-inspired convolutional spiking neural networks trained and tested on the CIFAR-10 dataset. The CIFAR-10 dataset was selected because the authors aim to use the target architecture in robotics applications and processing live images captured by RGB cameras. The proposed architecture incorporates 2 types of learning, namely, \textit{phase 1:} unsupervised STDP learning for training the synaptic connections between the L1 and L2a, L2a and L2b, and L2b and L2a; Inhibitory synapses are used to connect the neurons from L2b to L2a for providing lateral inhibition and \textit{phase 2:} ReSuMe is used for training the synaptic connections between the L2a and L3, and intra L3 neurons connectivity.
LIF neurons with adaptive threshold were used to inhibit neurons from spiking with very high spike rates when exposed to very high currents.
The CIFAR-10 pixel intensity values were normalised and the current constant $I_k$ was tuned for producing a spike rate proportional to the pixel intensity value. Also, the Monte Carlo algorithm was used for selecting the initial STDP weights for synapses connecting L2a and L3 neurons. Static synapses were used during the testing phase and the trained weights loaded into those synapses. Only the spike rate was used to select the correct class using the winner-takes-all. The NatCSNN was trained on the 50,000 training batch and tested on the 10,000 testing batch of the CIFAR-10 dataset. The NEST-simulator was used to implement the NatCSNN because, it is currently being used by the Neurorobotics platform for emulating bio-inspired neural networks. 
The main contributions of this paper are (a) a 3-layer bio-inspired convolutional neural network architecture designed to process natural images with no pre-processing required that exhibits a better accuracy than 3-layer classical CNN or ANN. (b)the use of combination of unsupervised learning for training the middle layers, (c)incorporation of lateral inhibition for reducing the background interference and (d) a flexible architecture that enables the possibility of processing live captured images (during test mode).
Future work, includes  expanding the current work to live-captured images and process such images on-the-fly. A more detailed analysis of the pre-trained weights and a pruning approach for synapses/neurons that have no influence over the simulation (weights not trained during the training phases) will be conducted. The adaptation of the NatCSNN to work with coloured images may include the use of hybrid approaches where the convolution layers could be replaced by efficient computer vision and deep learning approaches. Further testing under different test conditions (\textit{e.g.} different light conditions, different image size, more objects per scene, etc.) will also be conducted. Finally we intend to apply the NatCSNN in both robotic simulation (\textit{i.e.} Gazebo) and in real robotic applications. 

\bibliographystyle{splncs04}
\bibliography{references.bib}

\end{document}